\documentclass{article}

    \PassOptionsToPackage{numbers, compress}{natbib}
    \usepackage[preprint]{neurips_2021}

\usepackage[utf8]{inputenc} % allow utf-8 input
\usepackage[T1]{fontenc}    % use 8-bit T1 fonts
\usepackage{hyperref}       % hyperlinks
\usepackage{url}            % simple URL typesetting
\usepackage{booktabs}       % professional-quality tables
\usepackage{amsfonts}       % blackboard math symbols
\usepackage{nicefrac}       % compact symbols for 1/2, etc.
\usepackage{microtype}      % microtypography
\usepackage{xcolor}         % colors
\usepackage{subfigure}
\usepackage{multirow}
\usepackage{xspace}
\usepackage{listings}
\usepackage{xcolor}
\usepackage[linesnumbered,vlined,ruled,noend]{algorithm2e}
\usepackage{wrapfig}
\usepackage{enumitem}
\usepackage{CJKutf8}
\usepackage{floatrow}
\newfloatcommand{capbtabbox}{table}[][\FBwidth]
\usepackage{graphicx}
\usepackage{amsmath,amsfonts,graphicx}
\usepackage{amsthm}

\usepackage{babel}
\usepackage{caption}

\usepackage{makecell}
\usepackage{CJKutf8}

\title{Graph Attention MLP with Reliable Label Utilization}

\author{%
  Wentao Zhang$^{1}$, Ziqi Yin$^3$, Zeang Sheng$^1$, Wen Ouyang$^2$, Xiaosen Li$^2$ \\
  \textbf{Yangyu Tao$^2$, Zhi Yang$^1$, Bin Cui$^1$} \\
  $^1$EECS, Peking University $^2$Tencent Inc. $^3$Beijing Institute of Technology\\
}

\begin{document}

\maketitle

\begin{abstract}
Graph neural networks (GNNs) have recently achieved state-of-the-art performance in many graph-based applications.
Despite the high performance, they typically need to perform expensive recursive neighborhood expansions during every training epoch and has the scalability issue. 
Moreover, most of them are inflexible since they
are restricted to fixed-hop neighborhoods and insensitive to actual receptive field requirements for each node. 
We circumvent these limitations by introducing a scalable and flexible method: Graph Attention Multi-Layer Perceptron (GAMLP). 
Following the routine of decoupled GNNs, the feature propagation in GAMLP is executed during pre-computation, which helps it maintain high scalability.
With three proposed receptive field attention, each node in GAMLP is flexible in leveraging the propagated features over the different sizes of reception field. 
We conduct extensive evaluations on two large open graph benchmarks (ogbn-products and ogbn-papers100M), demonstrating that GAMLP not only achieves state-of-the-art performance, but also enjoys high scalability and efficiency.

\end{abstract}

\section{Introduction}
Graph Neural Networks (GNNs) are powerful deep neural networks for graph-structured data, becoming the \emph{de facto} method in many semi-supervised and unsupervised graph representation learning scenarios such as node
classification, link prediction, recommendation, and knowledge graphs~\cite{kipf2016semi, hamilton2017inductive, bo2020structural, cui2020adaptive, fan2019graph, trouillon2017knowledge}.
Through stacking $K$ graph convolution layers, GNNs can learn node representations by utilizing information from the $K$-hop neighborhood and thus enhance the model performance by getting more unlabeled nodes involved in the training process.

Unlike the images, text, or tabular data, where training data are independently distributed, graph data contains extra relationship information between nodes.
Besides, the real-world graph is usually huge. For example, the users and their relationships in Wechat can be formed as a graph, and this graph has billions of nodes and ten billion edges. 
Every node in a $K$-layer GNN will incorporate a set of nodes, including the node itself and its $K$-hop neighbors.
This set is called Receptive Field(RF).
As the size of RF grows exponentially to the number of GNN layers, the rapidly expanding RF introduces high computation and memory cost in a single machine. 
Besides, even in a distributed environment, GNN has to read great amount of data of neighboring nodes to compute the single target node representation, leading to high communication cost. 
Despite their effectiveness, the utilization of neighborhood information in GNNs leads to the scalability issue for training on large graphs.

A commonly used approach to tackle this issue is sampling, such as node sampling~\cite{hamilton2017inductive, DBLP:conf/icml/ChenZS18}, layer sampling~\cite{DBLP:conf/nips/Huang0RH18,DBLP:conf/iclr/ChenMX18} and graph sampling~\cite{chiang2019cluster,DBLP:conf/iclr/ZengZSKP20}. However, the sampling-based methods are imperfect because they still face high communication costs, and the sampling quality highly influences the model performance. 
Besides, a recent direction for scalable GNNs is based on model simplification.
For example, Simplified GCN (SGC)~\cite{wu2019simplifying} decouples the feature propagation and the non-linear transformation process, and the former is executed during pre-processing.  
Unlike the sampling-based methods, which still need feature propagation in each training epoch, this time-consuming process in SGC is only executed once, and only the nodes of training set get involved in the model training.
As a result, SGC is computation and memory efficient in a single machine and scalable in distributed settings since it does not require fetch features of neighboring nodes in the model training process.
Despite the scalability, SGC adopts fixed layers of feature propagation, leading to a fixed RF of all nodes. 
Such graph-wise propagation lacks the flexibility to model the interesting correlations on node features under different reception fields. 
This either makes that long-range dependencies cannot be fully leveraged due to the undersized RF, or loses local information due to introducing over-smoothed noise with the oversized RF.
Both results in non-optimal discriminative node representations.

\begin{figure*}[tp!]
\centering  
\subfigure[Inconsistent optimal steps]{
\label{fig:ob1}
\scalebox{0.6}{
   \includegraphics[width=1\linewidth]{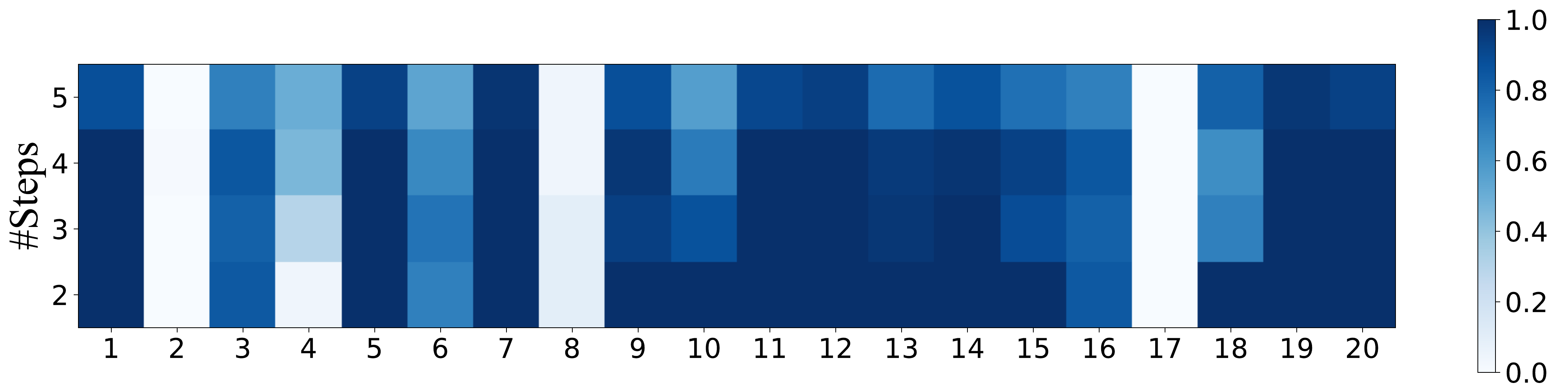}
 }}
\centering
\subfigure[Inconsistent RF expansion speed]{
\label{fig:ob2}
\scalebox{0.35}{
   \includegraphics[width=1\linewidth]{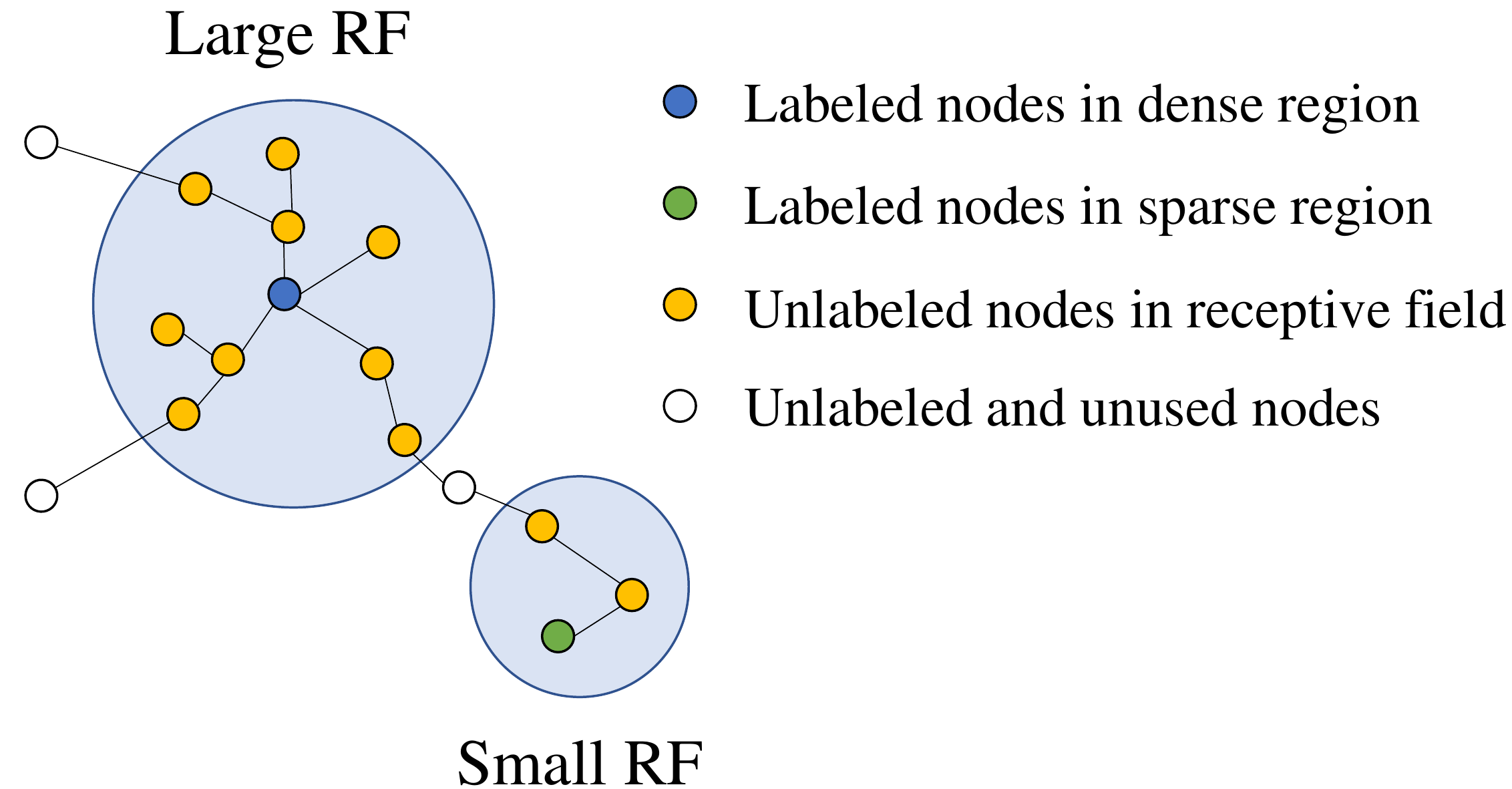}
 }}
\caption{  (Left) Test accuracy of SGC on 20 randomly sampled nodes of Citeseer. The X-axis is the node id, and Y-axis is the propagation steps (layers). The color from white to blue represents the ratio of being predicted correctly in 50 different runs. (Right) The local graph structures for two nodes in different regions; the node in the dense region has larger RF within two iterations of propagation.
}
\label{fig.observation}
\vspace{-2mm}
\end{figure*}

Lines of simplified models have been proposed to better use propagated features under different propagation layers and RF. 
As graph-wise propagation only considers features under a fixed layer, SIGN~\cite{frasca2020sign} proposes to concatenate all these features without information loss, while S$^2$GC~\cite{zhu2021simple} averages all these features to generate the combined feature with the same dimension. 
Although they have considered the influence of different layers, the importance of different propagated features is ignored. 
Under a large propagation layer, some over-smoothed features with oversized RF will introduce feature noise and degrade the model performance.
GBP~\cite{chen2020scalable} tackles this issue by adopting a constant decay factor for weighted average in the propagated features. 
Motivated by Personalized PageRank, the propagated features with larger propagation layers face a higher risk of over-smoothing, and they will contribute less to the final averaged features in GBP. 
All these methods adopt a layer-wise propagation mechanism and consider the features after different layers of propagation.
Despite their effectiveness, they fail to consider the feature combination from a node-wise level.

As shown in Figure~\ref{fig:ob1}, different nodes require different propagation steps and corresponding smoothness levels. Besides, the homogeneous and non-adaptive feature averaging may be unsuitable for all nodes due to the inconsistent RF expansion speed shown in Figure~\ref{fig:ob2}.
To support scalable and node-adaptive graph learning, we propose a novel MLP with three RF attention, abbreviated as GAMLP.
Experimental results demonstrate that GAMLP achieves the state-of-the-art performance on the three largest ogbn datasets, while maintains high scalability and efficiency.

\section{Preliminaries}
In this section, we introduce the notations
and review some current works tackling GNN scalability.

\textbf{Notations.} 
We consider an undirected graph $\mathcal{G}$ = ($\mathcal{V}$,$\mathcal{E}$) with $|\mathcal{V}| = N$ nodes and $|\mathcal{E}| = M$ edges. We denote by $\mathbf{A}$ the adjacency matrix of $\mathcal{G}$, weighted or not. Nodes can possibly have features vector of size $d$, stacked up in an $N \times d$ matrix $\mathbf{X}$. $\mathbf{D}=\operatorname{diag}\left(d_{1}, d_{2}, \cdots, d_{N}\right) \in \mathbb{R}^{N \times N}$ denotes the degree matrix of $\mathbf{A}$, where $d_{i}=\sum_{v_{j} \in \mathcal{V}} \mathbf{A}_{i j}$ is the degree of node $v_{i}$. 
Suppose $\mathcal{V}_l$ is the labeled set, and our goal is to predict the labels for nodes in the unlabeled set $\mathcal{V}_u$ with the supervision of $\mathcal{V}_l$.

\textbf{Sampling.} A commonly used method to tackle the scalability issue (i.e., the recursive neighborhood expansion) in GNN is sampling. As a node-wise sampling method, GraphSAGE~\cite{hamilton2017inductive} randomly samples a fixed size set of neighbors for computing in each mini-batch. VR-GCN~\cite{DBLP:conf/icml/ChenZS18} analyzes the variance reduction so that it can reduce the size of samples with an additional memory cost.
For the layer-wise sampling, Fast-GCN~\cite{DBLP:conf/iclr/ChenMX18} samples a fixed number of nodes at each layer, and ASGCN~\cite{DBLP:conf/nips/Huang0RH18} proposes the adaptive layer-wise sampling with better variance control.
In the graph level, Cluster-GCN~\cite{chiang2019cluster} firstly clusters the nodes and then samples the nodes in the clusters, and GraphSAINT~\cite{DBLP:conf/iclr/ZengZSKP20} directly samples a subgraph for mini-batch training. 
As an orthogonal way to model simplification, sampling has already been widely used in many GNNs and GNN systems~\cite{distdgl_ai3_2020, aligraph_vldb_2019, pygeometric_iclr_2019}. 

\textbf{Graph-wise Propagation.} 
Recently studies have observed that non-linear feature transformation contributes little to the performance of the GNNs as compared to feature propagation.
Thus, a new direction recently emerging for scalable GNN is based on the \emph{simplified} GCN (SGC)~\cite{wu2019simplifying}, which successively removes nonlinearities and collapsing weight matrices between consecutive layers. This reduces GNNs into a linear model operating on $K$-layers propagated features: 
\begin{small}
\begin{equation}
    \mathbf{X}^{(K)}=\mathbf{\hat{A}}^K \mathbf{X}^{(0)}, \qquad  \mathbf{Y} = \text{softmax}(\mathbf{\Theta} \mathbf{X}^{(K)}),
\end{equation}
\label{eq_GC}
\end{small}
\noindent where $\mathbf{X}^{(0)} =\mathbf{X}$, $\mathbf{X}^{(K)}$ is the $K$-layers propagated feature, and $\mathbf{\hat{A}} = \widetilde{\mathbf{D}}^{r-1}\widetilde{\mathbf{A}}\widetilde{\mathbf{D}}^{-r}$.
By setting $r = $ 0.5, 1 and 0, $\mathbf{\hat{A}}$ represents the symmetric normalization adjacency matrix  $\widetilde{\mathbf{D}}^{-1/2}\widetilde{\mathbf{A}}\widetilde{\mathbf{D}}^{-1/2}$~\cite{DBLP:conf/iclr/KlicperaBG19}, the transition probability matrix $\widetilde{\mathbf{A}}\widetilde{\mathbf{D}}^{-1}$~\cite{DBLP:conf/iclr/ZengZSKP20}, or the reverse transition probability matrix $\widetilde{\mathbf{D}}^{-1}\widetilde{\mathbf{A}}$~\cite{xu2018representation}, respectively. 
As the propagated features  $\mathbf{X}^{(K)}$ can be precomputed, SGC is more scalable and efficient for the large graph. However, such graph-wise propagation restricts the same propagation steps and a fixed RF for each node. Therefore, some nodes' features may be over-smoothed or under-smoothed due to the inconsistent RF expansion speed, leading to non-optimal performance.

\textbf{Layer-wise Propagation.} Following SGC, some recent methods adopt layer-wise propagation to combine the features with different propagation layers. SIGN~\cite{frasca2020sign} proposes to concatenate the different iterations of propagated features with linear transformation: $[\mathbf{X}^{(0)}\mathbf{W}_0, \mathbf{X}^{(1)}\mathbf{W}_1, ..., \mathbf{X}^{(K)}\mathbf{W}_K]$. S$^2$GC~\cite{zhu2021simple} proposes the simple spectral graph convolution to average the propagated features in different iterations as $\mathbf{X}^{(K)} = \sum \limits_{l=0}\limits^{K}\mathbf{\hat{A}}^l \mathbf{X}^{(0)}$. In addition, GBP~\cite{chen2020scalable} further improves the combination process by weighted averaging as $\mathbf{X}^{(K)} = \sum \limits_{l=0}\limits^{K} w_l\mathbf{\hat{A}}^l \mathbf{X}^{(0)}$ with the layer weight $w_l = \beta {(1-\beta)}^l$.
Similar to these works, we also use a linear model for higher training scalability. 
The difference lies in that we consider the propagation from a node-wise perspective and each node in GAMLP has a personalized combination of different steps of the propagated features.

\textbf{Self Supervision on GNN.}
Self supervision is widely used in GNN~\cite{DBLP:conf/icml/YouCWS20}, and the corresponding techniques can be mainly classified into the following categories: pretraining \& finetuning, self-training, self knowledge distillation, and multi-task learning. 

Following the paradigm of pretraining \& finetuning, some works~\cite{DBLP:conf/iclr/HuLGZLPL20, DBLP:conf/aaai/LuJ0S21, DBLP:conf/kdd/HuDWCS20} firstly train a GNN with the self-supervised task and then use the learned model parameters to initialize the other networks.
For example, ~\cite{DBLP:conf/iclr/HuLGZLPL20} pre-trains an expressive GNN at the level of individual nodes and avoids negative transfer, and improves generalization significantly across downstream tasks.
Besides, based on self training~\cite{triguero2015self}, some researches~\cite{DBLP:conf/aaai/LiHW18,sun2020multi} propose to assign “pseudo-labels”~\cite{lee2013pseudo} to the nodes with the high confident model prediction. As stated in previous works~\cite{bank2018improved, oliver2018realistic}, SLE~\cite{sun2021scalable} considers an unlabeled node confident if its corresponding class probability is larger than a predefined threshold. 
Different to the one hot label used in self training, recent works~\cite{zhang2020reliable, zhang2021rod, DBLP:conf/www/0002LS21} incorporate self-knowledge distillation into GNN and directly distill the model predicted soft label or embedding. Besides, we use a predefined threshold to filter reliable soft labels for distillation in this work.
At last, multi-task learning~\cite{ruder2017overview} regard the self-supervised task of GNN as a regularization term throughout the network training and thus help to improve the model generalizability~\cite{DBLP:conf/icml/YouCWS20,DBLP:journals/prl/ManessiR21}.

\textbf{Label Utilization on GNN.}
Labels of training nodes are conventionally only used as supervision signals in loss functions in most graph learning methods.
However, there also exist some graph learning methods that directly exploit the labels of training nodes.
Among them, the label propagation algorithm~\cite{zhu2002learnin} is the most well-known one.
It simply regards the partially observed label matrix $\mathbf{Y} \in \mathbb{R}^{N \times C}$ as input features for nodes in the graph and propagates the input features through the graph structure, where $C$ is the number of candidate classes.
UniMP~\cite{shi2020masked} proposes to map the partially observed label matrix $\mathbf{Y}$ to the dimension of the node feature matrix $\mathbf{X}$ and add these two matrices together as the new input feature.
To fight against the label leakage problem, UniMP further randomly masks the training nodes during every training epoch.

Instead of only using the hard training labels, Correct \& Smooth~\cite{huang2020combining} first trains a simple model such as an MLP and gets this simple model's predicted soft labels for unlabeled nodes. Then, it propagates the learning errors on the labeled nodes to connected nodes and smooths the output in a Personalized PageRank manner like APPNP~\cite{DBLP:conf/iclr/KlicperaBG19}. Besides, SLE~\cite{sun2021scalable} decouples the label utilization procedure in UniMP, and executes the propagation in advance. 
Unlike UniMP, ``label reuse''~\cite{wang2021bag}  concatenates the partially observed label matrix $\mathbf{Y}$ with the node feature matrix $\mathbf{X}$ to form the new input matrix.
Concretely,  it fills the missing elements in the partially observed label matrix $\mathbf{Y}$ with the soft label predicted by the model, and this newly generated $\mathbf{Y}'$ is again concatenated with $\mathbf{X}$ and then fed into the model to produce new predictions.

% After that, in the first Correct stage, C\&S propagates the learning errors on the labeled nodes (nodes in the training set) to connected nodes through graph structure.
% In the second Smooth stage, C\&S smooths the output of the first stage.
% To note that, during the propagation process in both stages, C\&S uses the personalized PageRank style transition probability matrix, which is the same as the one in APPNP~\cite{DBLP:conf/iclr/KlicperaBG19}. 
% SLE in SAGN+SLE~\cite{sun2021scalable} is another approach making use of label information.
% It decouples the label utilization procedure in UniMP, and executes the propagation in advance.
% SLE also adopts self-training, where the soft labels of high-confidence nodes predicted by the old model is also taken into consideration when executing propagation.
% The authors of~\cite{wang2021bag} propose ``label reuse'' to take advantage of the label information.
% Unlike UniMP, ``label reuse''~\cite{wang2021bag}  concatenates the partially observed label matrix $\mathbf{Y}$ with the node feature matrix $\mathbf{X}$ to form the new input matrix.
% During each training epoch, just after the predictions are generated, ``label reuse'' fills the missing elements in the partially observed label matrix $\mathbf{Y}$ with the soft label predicted by the model.
% This newly generated $\mathbf{Y}'$ is again concatenated with $\mathbf{X}$ and then fed into the model to produce new predictions.
% The number of ``label reuse'' iterations is a hyperparameter to tune.

\begin{figure}[tpb]
    \centering
    \includegraphics[width=0.9\textwidth]{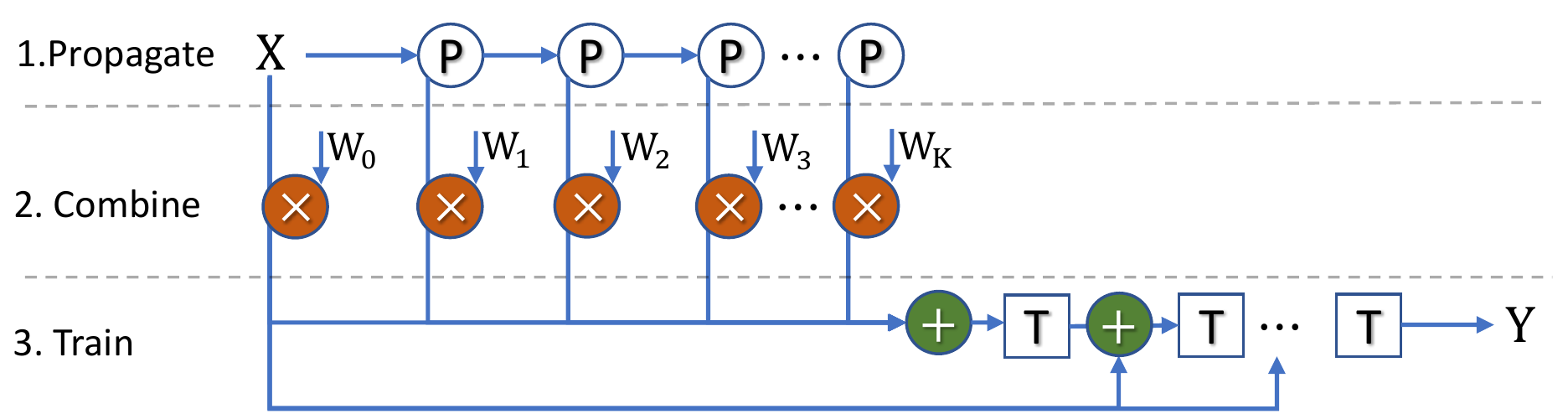}
    \caption{Overview of the proposed GAMLP, including (1) feature propagation, (2) feature combination with RF attention, and (3) MLP training. The feature propagation can be pre-processed.
}  
     \vspace{-2mm}
    \label{Fig.pipe}
\end{figure}

\section{The GAMLP Model}
\label{sec4}
\subsection{Overview}
As shown in Figure~\ref{Fig.pipe}, GAMLP decomposes the end-to-end GNN training into three parts: feature propagation, feature combination with RF attention, and the MLP training. 
As the feature propagation is pre-processed only once, and MLP training is efficient and salable, we can easily scale GAMLP to large graphs. Besides, with the RF attention, each node in GAMLP can adaptively get the suitable combination weights for propagated features under different RF, thus boosting model performance.

\begin{figure}[t]
    \vspace{-4mm}
	\centering
	\includegraphics[width=0.7\columnwidth]{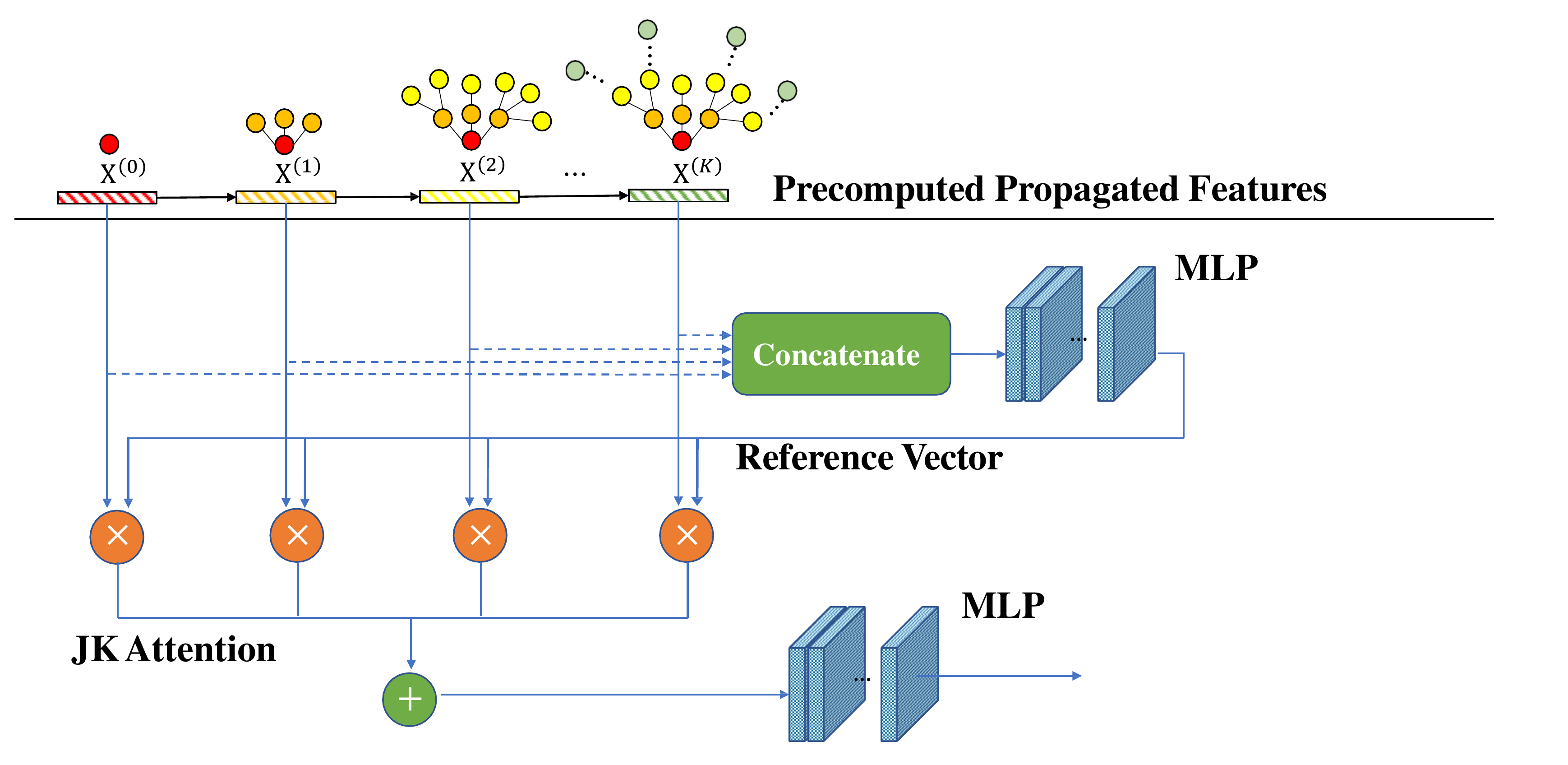}
	\caption{The architecture of GAMLP with JK Attention.}
	\label{fig:gmlp}
\end{figure}

\subsection{Establishment of GAMLP}

\subsubsection{Feature Propagation}  
We separate the essential operation of
GNNs — feature propagation by removing the neural network $\mathbf{\Theta}$ and
nonlinear activation $\delta$ for feature transformation. 
Specifically, we construct a parameter-free $K$-step feature propagation as:
\begin{equation}
    \mathbf{X}^{(l)} \gets \mathbf{T}\mathbf{X}^{(l-1)},\ \forall l=1,\ldots, K,
    \label{equ:prop}
\end{equation}
where $\mathbf{X}^{(l)}$ contains the features of a fixed RF: the node itself and its $l$-hop neighborhoods. 

After $K$-step feature propagation shown in E.q.~\ref{equ:prop}, we correspondingly get a list of propagated features under different propagation steps: $[\mathbf{X}^{(0)}, \mathbf{X}^{(1)}, \mathbf{X}^{(k)}, ..., \mathbf{X}^{(K)}]$.
For a node-wise propagation, we propose to average these propagated features in a weighted manner:
\begin{equation}
      \mathbf{H}=\sum_{l=0}^K \mathbf{W}_l \mathbf{X}^{(l)} ,
      \label{combine}
\end{equation}
where $\mathbf{W}_l  = Diag(\eta_{l})$ is the diagonal matrix derived from vector $\eta_{l}$, and $\eta_{l}$ is an $N$-dimension vector derived from vector $\eta_{l} = w_{i}(l), 1\le i\le N$, and $w_{i}(l)$ measures the importance of the $l$-step propagated feature for node $v_i$.
To satisfy different RF requirements for each node, we introduce three RF attention mechanisms to get $w_{i}(l)$.

\subsubsection{Receptive Field Attention}
\textbf{Smoothing Attention.} 
Suppose we execute feature propagation for infinite times. In that case, the node embedding within the same connected component will reach a stationary state, and it is hard to distinguish a specific node from others.
This issue is referred to as over-smoothing~\cite{li2018deeper}.
Concretely, when applying $\widetilde{\mathbf{D}}^{r-1}\tilde{\mathbf{A}}\widetilde{\mathbf{D}}^{-r}$ as adjacency matrix $\hat{\mathbf{A}}$, the stationary state follows
\begin{equation}
\label{stationary}
\hat{\mathbf{A}}^{\infty}_{i,j}  =  \frac{(d_i+1)^r(d_j+1)^{1-r}}{2m+n},
\end{equation}

To avoid the over-smoothing issue introduced by large RF, the weight $w_{i}(k)$ parameterized by node $i$ and aggregation step $k$ is defined as:
\begin{equation}
\label{iw}
 \widetilde{\mathbf{X}}_i^{(l)} = \mathbf{X}_i^{(l)} \parallel \mathbf{X}_i^{(\infty)}, \quad    {\widetilde{w}_{i}(l)} = \delta(\widetilde{\mathbf{X}}_i^{(l)} \cdot s), \quad  w_{i}(l) = {e^{\widetilde{w}_{i}(l)}}/{\sum \limits_{k=0}\limits^{K} e^{\widetilde{w}_{i}(k)}},
\end{equation}
where $\parallel$ stands for concatenation, and $s \in \mathbb{R}^{1 \times d}$ is a trainable vector.  Larger $w_{i}(l)$ means that the $l$-step propagated feature of node $v_i$ is more distant from the stationary state, and has less risk of being noise. Therefore, propagated feature with larger $w_{i}(k)$ should contribute more to the feature combination.

\textbf{Recursive Attention} 
At each propagation step $l$, suppose $s \in \mathbb{R}^{d}$ is a learnable parameter vector, we recursively measure the feature information gain compared with the previous combined feature as:
\begin{equation}
\small
\label{iw1}
\widetilde{\mathbf{X}}_i^{(l)} = \mathbf{X}_i^{(l)} \parallel \sum_{k=0}^{l-1} w_i(k) \mathbf{X}_i^{(k)},  \quad  w_{i}(k) = {e^{\widetilde{w}_{i}(k)}}/{\sum \limits_{j=0}\limits^{l-1} e^{\widetilde{w}_{i}(j)}},
\end{equation}
As $\widetilde{\mathbf{X}}_i^{(l-1)}$ combines the graph information under different propagation steps and RF,  large proportion of the information in $\widetilde{\mathbf{X}}_i^{(l)}$ may have already existed in $\sum_{k=0}^{l-1} w_i(k) \mathbf{X}_i^{(k)}$, leading to small information gain .
larger $w_{i}(l)$ means the feature $\mathbf{X}_i^{(l)}$ is more important to the current state of node $v_i$ since combining $\widetilde{\mathbf{X}}_i^{(l)}$ will introduce higher information gain.

\textbf{JK Attention} Jumping Knowledge Network (JK-Net)~\cite{xu2018representation} adopts layer aggregation to combine the node embeddings of different GCN layers, and thus it can leverage the propagated nodes information with different RF. Motivated by JK-Net, we propose to guide the feature combination process with the model prediction trained on all the propagated features.   
Figure \ref{fig:gmlp} shows the corresponding model architecture of GAMLP with JK attention, which includes two branches: the concatenated JK branch and the attention-based combination branch. We define the MLP prediction of the JK branch as $ \mathbf{E}_i = \text{MLP}(\mathbf{X}_i^{(1)}\parallel\mathbf{X}_i^{(2)}\parallel ... \parallel\mathbf{X}_i^{(K)})$, and then define the combination weight as:
\begin{equation}
\label{iw}
\widetilde{\mathbf{X}}_i^{(l)} = \mathbf{X}_i^{(l)} \parallel \mathbf{E}_i,  \quad   {\widetilde{w}_{i}(l)} = \delta(\widetilde{\mathbf{X}}_i^{(l)} \cdot s), \quad w_{i}(l) = {e^{\widetilde{w}_{i}(l)}}/{\sum \limits_{k=0}\limits^{K} e^{\widetilde{w}_{i}(k)}},
\end{equation}

The JK branch aims to create a multi-scale feature representation for each node, which helps the attention mechanism learn the weight $w_i(k)$.
%to recognize the correlation among node representations of different propagation steps.
The learned weights are then fed into the attention-based combination branch to generate each node's refined attention feature representation.
As the training process continues, the attention-based combination branch will gradually 
%remove the noisy level of localities and 
emphasize those neighborhood regions that are more helpful to the target nodes.
% \red{why capable of modeling a wider neighborhood while enhancing correlations?}
The JK attention can model a wider neighborhood while enhancing correlations, bringing a better feature representation for each node.

\subsubsection{Incorporating Label Propagation.}
%\textbf{label propagation}
%To reinforce the model performance, we propose a simple and scalable way to take advantage of the node labels of the training set. 
Similar to previous works~\cite{shi2020masked, huang2020combining, sun2021scalable, wang2021bag}, we also use label propagation for better utilization of label information .
Concretely, the label embedding matrix $\hat{\mathbf{Y}}^{(0)}$ is firstly initialized as all zero.
Then, we use the hard training labels, $\mathbf{Y}_{\mathcal{V}_l}$, to fill in the all zero matrix and propagate it with the normalized adjacency matrix $\hat{\mathbf{A}}$:
\begin{small}
\begin{equation}
    \hat{\mathbf{Y}}_{\mathcal{V}_l}^{(0)} = \mathbf{Y}_{\mathcal{V}_l}, \hat{\mathbf{Y}}^{(k+1)} = \hat{\mathbf{A}} \hat{\mathbf{Y}}^{(k)},
\end{equation}
\end{small}
where $\mathcal{V}_l$ is the labeled node set.
After $K$-steps Label Propagation, we get the final label embedding  $\hat{\mathbf{Y}}^{(K)}$ and then use it to enhance the model prediction.

% is fed into a simple MLP to map the embedding to the same space as the model prediction:
% \begin{equation}
%     \widetilde{\mathbf{Y}} =  \text{MLP}(\hat{\mathbf{Y}}^{(K)}),
% \end{equation}
% where 

\subsubsection{Model Training}
Previous work~\cite{zhang2021evaluating} shows that the main limitations of deep GNNs are the \textit{over-smoothing} introduced by large steps of propagation and \textit{model degradation} introduced by too many layers of non-linear transformation. The proposed attention-based feature propagation can adaptively leverage the propagated features over the different sizes of reception field and avoid the \textit{over-smoothing} issue. 

Large graph~\cite{zhang2021evaluating} require large steps of non-liner transformation. To tackle the \textit{model degradation} problem, we propose to use initial residual as:
\begin{equation}
    \hat{\mathbf{H}}^{(l)} \gets \delta(\mathbf{W}^{(l)}\hat{\mathbf{H}}^{(l-1)} + \mathbf{X}^{(0)}),\ l=1,\ldots, L,
    \label{equ:prop}
\end{equation}
where $L$ is the MLP layers, and $\mathbf{H}^{(0)} = \hat{\mathbf{H}}$ is the combined feature matrix. 

Then the output of this $L$ layers MLP is added with the label embedding from Sec.~3.2.3 to get the final output embedding:
\begin{equation}
    \widetilde{\mathbf{H}} = \hat{\mathbf{H}}^{(L)} + \text{MLP}(\hat{\mathbf{Y}}^{(K)}).
\end{equation}

The MLP here is used to map the embedding $\hat{\mathbf{Y}}^{(K)}$ to the same space as $\hat{\mathbf{H}}^{(L)}$.

We adopt the Cross-Entropy (CE) measurement between the predicted softmax outputs and the one-hot ground-truth label distributions as the objective function: 
\begin{equation}
\begin{aligned}
      & \mathcal{L}_{CE}= -\sum_{i \in \mathcal{V}_l } \sum_{j} \mathbf{Y}_{ij} \log(\text{softmax}(\widetilde{\mathbf{H}})_{ij})
      \label{classi}
\end{aligned}
\end{equation}
where $\mathbf{Y}_i$ is the one-hot label indicator vector.

\subsubsection{Reliable Label Utilization (RLU)}
\paragraph{Reliable Label Propagation.}
To better utilize the predicted soft label (i.e., softmax outputs), we split the whole training process into multiple stages, each containing a full training procedure of the GAMLP model.
At the first stage, the GAMLP model is trained according to the above-mentioned procedure.
However, at later stages, we take advantage of the predicted reliable soft label of the last stage to improve the label embedding $\hat{\mathbf{Y}}^{(0)}$.
Here, we denote the prediction results of $m$-th stage as $\mathbf{P}^{(m)}$:
\begin{equation}
    \mathbf{P}^{(m)} = \text{softmax}(\hat{\mathbf{H}}/T), \quad T \in (0, 1],
\end{equation}
where the parameter $T$ controls the softness of the softmax distribution. 
Lower values of $T$ leads to more hardened distribution. 

Suppose we are now at the beginning of the $m$-th stage ($m>1$) of the whole training process.
Rather than just using the training labels to construct the initial label embedding $\hat{\mathbf{Y}}^{(0)}$, we adopt the predicted results for the nodes in the validation set and the test set at the last stage as well.
To ensure the reliability of the predicted soft label, we use a threshold $\epsilon$ to filter out the low-confident nodes in the validation set and the test set.
The formulation for the reliable label is as follows:
\begin{small}
\begin{equation}
\hat{\mathbf{Y}}^{(0)}_i= 
\left\{
\begin{array}{lll}
\mathbf{Y}_i, \ \ \text{if} \ \ i \in \mathcal{V}_l, \\
\mathbf{P}^{(m-1)}_i \ \ \text{if} \ \ i \in \mathcal{V}_r, \\
\mathbf{0}, \ \ \text{otherwise}.
\end{array}
\right.
\hat{\mathbf{Y}}^{(k+1)} = \hat{\mathbf{A}} \hat{\mathbf{Y}}^{(k)}.
\end{equation}
\end{small}
In the formulation, the reliable node set $\mathcal{V}_r$ is composed of nodes whose predicted probability belonging to the most likely class at $(m-1)$-th stage is greater than the threshold $\epsilon$.
%In the formulation, the reliable node-set $\mathcal{V}_r$ is composed of nodes whose confidence score $\alpha_i$ is greater than the threshold $\epsilon$, where $w_i$ is the predicted probability  belonging to the most likely class at $(m-1)$-th stage .

\paragraph{Reliable Label Distillation.}
To fully take advantage of the helpful information of the last stage, we also included a knowledge distillation module in our model.
Again to guarantee the reliability of the knowledge distillation module, we only include the nodes in the reliable node set $\mathcal{V}_r$ at $m$-th stage ($m>1$) and then define the weighted KL divergence as:
\begin{small}
\begin{equation}
\begin{aligned}
      & \mathcal{L}_{KD} =  \sum_{i \in \mathcal{V}_{r}}\sum_{j}\alpha_i\mathbf{P}_{ij}^{(m-1)} \log \frac{\mathbf{P}_{ij}^{(m-1)}}{\mathbf{P}_{ij}^{(m)}},
\end{aligned}
\end{equation}
\end{small}
where $\alpha_i$ for reliable node $i$ is its predicted probability belonging to the mostly likely class at $(m-1)$-th stage.
We further incorporate $\alpha_i$ here to better guide the distillation process, assigning higher weights to more confident nodes.

The complete training loss for $m$-th stage ($m>1$) is defined as:
\begin{small}
\begin{equation}
    \mathcal{L} = \mathcal{L}_{CE} + \gamma \mathcal{L}_{KD},
\end{equation}
\end{small}
where $\gamma$ is a hyperparameter, balancing the importance of the knowledge distillation module.

\subsection{Relation with current methods}
\textbf{GAMLP vs. GBP.}
Both GAMLP and GBP propose to weighted average the propagated features under different propagation steps and RF. However, GBP adopts a layer-wise propagation and ignores the inconsistent RF expansion speed for different nodes. As the optimal propagation steps and smoothing levels of different nodes are different, some nodes may face the over-smoothing or under-smoothing issue even propagated the same step. GAMLP considers the feature propagation in a more fine-grained node perspective. Compared with GBP, SGC, and S$^2$GC, the limitation of GAMLP is that all the propagated features are required in the model training, leading to high memory cost.

\textbf{GAMLP vs. GAT.}
Each node in a GAT layer learns to weighted combine the embedding (or feature) of its neighborhoods with an attention mechanism, and the attention weights are measured by the local information in a fixed RF -- the node itself and its direct neighbors. 
Different from the attention mechanism in GAT, GAMLP considers more global information under different RF. 

\textbf{GAMLP vs. DAGNN.}
Similar to AP-GCN~\cite{spinelli2020adaptive}, DAGNN can adaptively learn the combination weights via the gating mechanism and thus assign proper weights for different nodes. 
However, the gating mechanism in DAGNN is correlated with the parameterized node embedding rather than the training-free node feature used in GAMLP, leading to low scalability and efficiency. 

\textbf{GAMLP vs. JK-Net.}
Motivated by JK-Net, GAMLP with JK attention concatenate the propagated features under different propagation steps. However, the model prediction based on the concatenated feature is just used as a reference vector for the attention-based combination branch in GAMLP rather than the final results. Compared with JK-Net, GAMLP with JK attention is more effective in alleviating the over-smoothing and scalability issue that deep architecture introduces. 

\textbf{GAMLP vs. SAGN.}
Both GAMLP and SAGN propose to do node specific propagation in GNN. Specially, they enable the adaptive propagation in a feature message passing way~\cite{zhang2021gmlp} and thus get high scalability. More concretely, SAGN learns the node-specific attention weights with the original node feature. Based on the observation of inconsistent RF expansion speed, GAMLP adopts three receptive field attention mechanisms to better combine the propagated features over different sizes of reception fields. Besides, the initial residual connection~\cite{chen2020simple} is also employed in GAMLP to support large steps of non-linear transformation.

\section{Experiments}
% \zwt{1. 数据集补充: cora，citeseer，PubMed，Reddit，PPI \\ 2. 模型补充：GMALP不带RLU的}

In this section, we verify the effectiveness of GAMLP on seven graph datasets. 
We aim to answer the following two questions. \textbf{Q1:} Compared with current methods, can GAMLP achieve higher predictive accuracy?
\iffalse
\textbf{Q2:} Can GAMLP  deal with the over-smoothing issue introduced by large propagation depth? \textbf{Q3:} What makes GAMLP effective?
\fi
\textbf{Q2:} Why GAMLP is effective? 

\subsection{Experimental Setup}

\begin{table*}[tpb!]
\small
\centering
\caption{Overview of datasets.}
\label{Datasets}
\resizebox{.95\linewidth}{!}{
\begin{tabular}{cccccc}
\toprule
\textbf{Dataset}&\textbf{\#Nodes}& \textbf{\#Features}&\textbf{\#Edges}&\textbf{\#Classes}&\textbf{\#Train/Val/Test}\\
\midrule
%Cora& 2,708 & 1,433 &5,429&7& 140/500/1,000\\
%Citeseer& 3,327 & 3,703&4,732&6& 120/500/1,000\\
%PubMed& 19,717 & 500 &44,338&3& 60/500/1,000\\
%ogbn-mag& 1,939,743 & 128 & 21,111,007 & 349 &  626K/66K/44K & citation network\\
ogbn-products& 2,449,029 & 100 & 61,859,140 & 47 & 196K/49K/2,204K \\
ogbn-papers100M & 111,059,956 & 128 & 1,615,685,872 & 172 & 1,207K/125K/214K\\
%Industry & 1,000,000 & 64 & 1,434,382 & 253 & 5K/10K/30K&short-form video network\\
\bottomrule
\end{tabular}}
\end{table*}

\textbf{Datasets and baselines.}
%We conduct the experiments on three commonly used citation networks (Citeseer, Cora, and PubMed) in~\cite{kipf2016semi},  (2) and the ogbn-products and ogbn-papers100M datasets in~\cite{hu2021ogb}.
We conduct the experiments on the ogbn-products and ogbn-papers100M datasets in~\cite{hu2021ogb}.
The dataset statistics are shown in Table~\ref{Datasets}.
For the comparison on the ogbn-products dataset, we choose the following baseline methods: GCN~\cite{kipf2016semi}, GraphSAGE~\cite{hamilton2017inductive}, SIGN~\cite{frasca2020sign}, DeeperGCN~\cite{li2020deepergcn}, SAGN+0-SLE and SAGN+2-SLE~\cite{sun2021scalable}, UniMP~\cite{shi2020masked}, and MLP+C\&S~\cite{huang2020combining}.
For the comparison on the ogbn-papers100M dataset, we choose the following baseline methods: SGC~\cite{wu2019simplifying}, SIGN and SIGN-XL~\cite{frasca2020sign}, SAGN+0-SLE and SAGN+2-SLE~\cite{sun2021scalable}.
The validation and test accuracy of these all the baseline methods are directly from the OGB leaderboard.

\textbf{Implementation.}

To alleviate the influence of randomness, we repeat each method ten times and report the mean performance and the standard deviations.
The experiments are conducted on a machine with Intel(R) Xeon(R) Platinum 8255C CPU@2.50GHz, and a single Tesla V100 GPU with 32GB GPU memory. 
The operating system of the machine is Ubuntu 16.04. 
As for software versions, we use Python 3.6, Pytorch 1.7.1, and CUDA 10.1.
The hyper-parameters in each baseline are set according to the original paper if available. Please refer to Appendix B for the detailed hyperparameter settings for our GAMLP+RLU.

\begin{table*}[tpb!]
\caption{Test accuracy on ogbn-products dataset.}
%\vspace{-2mm}
\centering
{
\noindent
\renewcommand{\multirowsetup}{\centering}
\resizebox{0.75\linewidth}{!}{
\begin{tabular}{c|cc}
\toprule
\textbf{Methods} & \textbf{Validation Accuracy} & \textbf{Test Accuracy} \\
\midrule
GCN & 92.00$\pm$0.03 & 75.64$\pm$0.21 \\
GraphSAGE & 92.24$\pm$0.07 & 78.50$\pm$0.14 \\
SIGN & 92.99$\pm$0.04 & 80.52$\pm$0.16 \\
DeeperGCN & 92.38$\pm$0.09 & 80.98$\pm$0.20 \\
UniMP & 93.08$\pm$0.17 & 82.56$\pm$0.31 \\
SAGN+0-SLE & 93.27$\pm$0.04 & 83.29$\pm$0.18 \\
MLP+C\&S & 91.47$\pm$0.09 & 84.18$\pm$0.07 \\
SAGN+2-SLE & 92.87$\pm$0.03 & \underline{84.28$\pm$0.14} \\
%\textbf{GAMLP} & - & -  \\
\textbf{GAMLP} & 93.12$\pm$0.03 & 83.54$\pm$0.09  \\
\textbf{GAMLP+RLU} & 93.24$\pm$0.05 & \textbf{84.59$\pm$0.10}  \\
\bottomrule
\end{tabular}}}
\label{table.products_performance}
%\vspace{-3mm}
\end{table*}

\begin{table*}[tpb!]
\caption{Test accuracy on ogbn-papers100M dataset.}
\vspace{-2mm}
\centering
{
\noindent
\renewcommand{\multirowsetup}{\centering}
\resizebox{0.75\linewidth}{!}{
\begin{tabular}{c|cc}
\toprule
\textbf{Methods} & \textbf{Validation Accuracy} & \textbf{Test Accuracy} \\
\midrule
SGC & 66.48$\pm$0.20 & 63.29$\pm$0.19 \\
SIGN & 69.32$\pm$0.06 & 65.68$\pm$0.06 \\
SIGN-XL & 69.84$\pm$0.06 & 66.06$\pm$0.19 \\
SAGN+0-SLE & 71.06$\pm$0.08 & 67.55$\pm$0.15 \\
SAGN+2-SLE & 71.31$\pm$0.10 & \underline{68.00$\pm$0.15} \\
%\textbf{GAMLP} & - & -  \\
\textbf{GAMLP} & 71.17$\pm$0.14 & 67.71$\pm$0.20  \\
\textbf{GAMLP+RLU} & 71.59$\pm$0.05 & \textbf{68.25$\pm$0.11}  \\
\bottomrule
\end{tabular}}}
\label{table.100m_performance}
\vspace{-3mm}
\end{table*}

\subsection{Experimental Results.}  

\textbf{End-to-end comparison.}  To answer \textbf{Q1}, Table~\ref{table.products_performance},~\ref{table.100m_performance} show the validation and test accuracy of our GAMLP and GAMLP+RLU and all the baseline methods.
The raw performance of our method GAMLP is very competitive compared to other baseline methods, which outperforms the current SOTA single model UniMP and SAGN+0-SLE.
With the help of RLU, the test accuracy of GAMLP+RLU gets a decent improvement further and outperforms all the compared baselines, exceeding the strongest baseline SAGN+2-SLE by 0.17\% and 0.25\% on ogbn-products and ogbn-papers100M datasets, respectively.
The performance of our GAMLP empirically illustrates the effectiveness of our proposed attention mechanisms.

  \begin{figure*}[tpb!]
    \centering
    \includegraphics[width=0.55\linewidth]{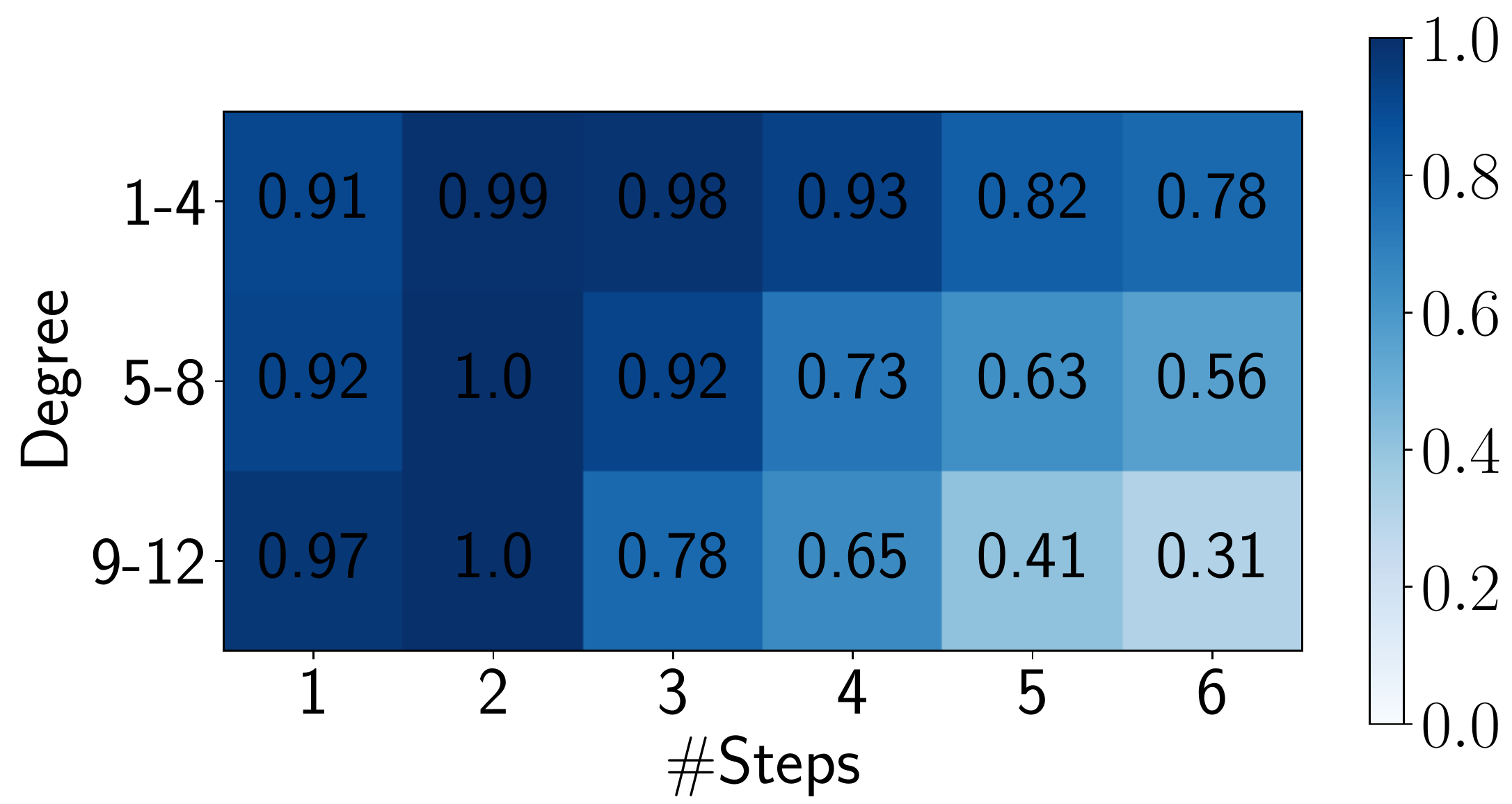}
    % \vspace{-4mm}
    \caption{The average attention weights of propagated features of different steps on 60 randomly selected nodes from ogbn-products. }
    % \vspace{-4mm}
    \label{interpretability}
\end{figure*}

\textbf{Interpretability.} 
GAMLP can adaptively and effectively combine multi-scale propagated features for each node.
To demonstrate this, Figure~\ref{interpretability} shows the average attention weights of propagated features according to the number of steps and degrees of input nodes, where the maximum step is 6. In this experiment, we randomly select 20 nodes for each degree range (1-4, 5-8, 9-12) and plot the relative weight based on the maximum value. 
We get two observations from the heat map: 1) The 1-step and 2-step propagated features are always of great importance, which shows that GAMLP captures the local information as those widely 2-layer methods do; 2) The weights of propagated features with larger steps drop faster as the degree grows, which indicates that our attention mechanism could prevent high-degree nodes from including excessive irrelevant nodes which lead to over-smoothing. From the two observations, we conclude that GAMLP is able to identify the different RF demands of nodes and explicitly weight each propagated feature.

\section{Conclusion}
This paper presents Graph Attention Multilayer Perceptron (GAMLP), a scalable, efficient, and powerful graph learning method based on the reception field attention. Concretely, GAMLP defines three principled attention mechanisms, i.e., smoothing attention, recursive attention, and JK attention, and each node in GAMLP can leverage the propagated features over different size of RF in a node-specific way. Extensive experiments on three large ogbn graphs verify the effectiveness of the proposed method.

%%%%%%%%%%%%%%%%%%%%%%%%%%%%%%%%%%%%%%%%%%%%%%%%%%%%%%%%%%%%
\bibliographystyle{abbrv}
\bibliography{reference}

\begin{thebibliography}{10}

\bibitem{bank2018improved}
D.~Bank, D.~Greenfeld, and G.~Hyams.
\newblock Improved training for self training by confidence assessments.
\newblock In {\em Science and Information Conference}, pages 163--173.
  Springer, 2018.

\bibitem{bo2020structural}
D.~Bo, X.~Wang, C.~Shi, M.~Zhu, E.~Lu, and P.~Cui.
\newblock Structural deep clustering network.
\newblock In {\em Proceedings of The Web Conference 2020}, pages 1400--1410,
  2020.

\bibitem{DBLP:conf/iclr/ChenMX18}
J.~Chen, T.~Ma, and C.~Xiao.
\newblock Fastgcn: Fast learning with graph convolutional networks via
  importance sampling.
\newblock In {\em 6th International Conference on Learning Representations,
  {ICLR} 2018, Vancouver, BC, Canada, April 30 - May 3, 2018, Conference Track
  Proceedings}, 2018.

\bibitem{DBLP:conf/icml/ChenZS18}
J.~Chen, J.~Zhu, and L.~Song.
\newblock Stochastic training of graph convolutional networks with variance
  reduction.
\newblock In {\em Proceedings of the 35th International Conference on Machine
  Learning, {ICML} 2018, Stockholmsm{\"{a}}ssan, Stockholm, Sweden, July 10-15,
  2018}, pages 941--949, 2018.

\bibitem{chen2020scalable}
M.~Chen, Z.~Wei, B.~Ding, Y.~Li, Y.~Yuan, X.~Du, and J.-R. Wen.
\newblock Scalable graph neural networks via bidirectional propagation.
\newblock {\em arXiv preprint arXiv:2010.15421}, 2020.

\bibitem{chen2020simple}
M.~Chen, Z.~Wei, Z.~Huang, B.~Ding, and Y.~Li.
\newblock Simple and deep graph convolutional networks.
\newblock In {\em International Conference on Machine Learning}, pages
  1725--1735. PMLR, 2020.

\bibitem{chiang2019cluster}
W.-L. Chiang, X.~Liu, S.~Si, Y.~Li, S.~Bengio, and C.-J. Hsieh.
\newblock Cluster-gcn: An efficient algorithm for training deep and large graph
  convolutional networks.
\newblock In {\em SIGKDD}, pages 257--266, 2019.

\bibitem{cui2020adaptive}
G.~Cui, J.~Zhou, C.~Yang, and Z.~Liu.
\newblock Adaptive graph encoder for attributed graph embedding.
\newblock In {\em SIGKDD}, pages 976--985, 2020.

\bibitem{fan2019graph}
W.~Fan, Y.~Ma, Q.~Li, Y.~He, E.~Zhao, J.~Tang, and D.~Yin.
\newblock Graph neural networks for social recommendation.
\newblock In {\em The World Wide Web Conference}, pages 417--426, 2019.

\bibitem{pygeometric_iclr_2019}
M.~Fey and J.~E. Lenssen.
\newblock Fast graph representation learning with {PyTorch Geometric}.
\newblock In {\em ICLR 2019 Workshop on Representation Learning on Graphs and
  Manifolds}, 2019.

\bibitem{frasca2020sign}
F.~Frasca, E.~Rossi, D.~Eynard, B.~Chamberlain, M.~Bronstein, and F.~Monti.
\newblock Sign: Scalable inception graph neural networks.
\newblock {\em arXiv preprint arXiv:2004.11198}, 2020.

\bibitem{hamilton2017inductive}
W.~L. Hamilton, R.~Ying, and J.~Leskovec.
\newblock Inductive representation learning on large graphs.
\newblock In {\em NeurIPS}, pages 1025--1035, 2017.

\bibitem{hu2021ogb}
W.~Hu, M.~Fey, H.~Ren, M.~Nakata, Y.~Dong, and J.~Leskovec.
\newblock Ogb-lsc: A large-scale challenge for machine learning on graphs.
\newblock {\em arXiv preprint arXiv:2103.09430}, 2021.

\bibitem{DBLP:conf/iclr/HuLGZLPL20}
W.~Hu, B.~Liu, J.~Gomes, M.~Zitnik, P.~Liang, V.~S. Pande, and J.~Leskovec.
\newblock Strategies for pre-training graph neural networks.
\newblock In {\em 8th International Conference on Learning Representations,
  {ICLR} 2020, Addis Ababa, Ethiopia, April 26-30, 2020}, 2020.

\bibitem{DBLP:conf/kdd/HuDWCS20}
Z.~Hu, Y.~Dong, K.~Wang, K.~Chang, and Y.~Sun.
\newblock {GPT-GNN:} generative pre-training of graph neural networks.
\newblock In {\em {KDD} '20: The 26th {ACM} {SIGKDD} Conference on Knowledge
  Discovery and Data Mining, Virtual Event, CA, USA, August 23-27, 2020}, pages
  1857--1867, 2020.

\bibitem{hu2020heterogeneous}
Z.~Hu, Y.~Dong, K.~Wang, and Y.~Sun.
\newblock Heterogeneous graph transformer.
\newblock In {\em Proceedings of The Web Conference 2020}, pages 2704--2710,
  2020.

\bibitem{huang2020combining}
Q.~Huang, H.~He, A.~Singh, S.-N. Lim, and A.~R. Benson.
\newblock Combining label propagation and simple models out-performs graph
  neural networks.
\newblock {\em arXiv preprint arXiv:2010.13993}, 2020.

\bibitem{DBLP:conf/nips/Huang0RH18}
W.~Huang, T.~Zhang, Y.~Rong, and J.~Huang.
\newblock Adaptive sampling towards fast graph representation learning.
\newblock In {\em Advances in Neural Information Processing Systems 31: Annual
  Conference on Neural Information Processing Systems 2018, NeurIPS 2018,
  December 3-8, 2018, Montr{\'{e}}al, Canada}, pages 4563--4572, 2018.

\bibitem{kipf2016semi}
T.~N. Kipf and M.~Welling.
\newblock Semi-supervised classification with graph convolutional networks.
\newblock {\em arXiv preprint arXiv:1609.02907}, 2016.

\bibitem{DBLP:conf/iclr/KlicperaBG19}
J.~Klicpera, A.~Bojchevski, and S.~G{\"{u}}nnemann.
\newblock Predict then propagate: Graph neural networks meet personalized
  pagerank.
\newblock In {\em 7th International Conference on Learning Representations,
  {ICLR} 2019, New Orleans, LA, USA, May 6-9, 2019}, 2019.

\bibitem{lee2013pseudo}
D.-H. Lee.
\newblock Pseudo-label: The simple and efficient semi-supervised learning
  method for deep neural networks.
\newblock In {\em Workshop on challenges in representation learning, ICML},
  volume~3, page 896, 2013.

\bibitem{li2020deepergcn}
G.~Li, C.~Xiong, A.~Thabet, and B.~Ghanem.
\newblock Deepergcn: All you need to train deeper gcns.
\newblock {\em arXiv preprint arXiv:2006.07739}, 2020.

\bibitem{DBLP:conf/aaai/LiHW18}
Q.~Li, Z.~Han, and X.~Wu.
\newblock Deeper insights into graph convolutional networks for semi-supervised
  learning.
\newblock In {\em Proceedings of the Thirty-Second {AAAI} Conference on
  Artificial Intelligence, (AAAI-18), the 30th innovative Applications of
  Artificial Intelligence (IAAI-18), and the 8th {AAAI} Symposium on
  Educational Advances in Artificial Intelligence (EAAI-18), New Orleans,
  Louisiana, USA, February 2-7, 2018}, pages 3538--3545, 2018.

\bibitem{li2018deeper}
Q.~Li, Z.~Han, and X.-M. Wu.
\newblock Deeper insights into graph convolutional networks for semi-supervised
  learning.
\newblock In {\em Proceedings of the AAAI Conference on Artificial
  Intelligence}, volume~32, 2018.

\bibitem{DBLP:conf/aaai/LuJ0S21}
Y.~Lu, X.~Jiang, Y.~Fang, and C.~Shi.
\newblock Learning to pre-train graph neural networks.
\newblock In {\em Thirty-Fifth {AAAI} Conference on Artificial Intelligence,
  {AAAI} 2021, Thirty-Third Conference on Innovative Applications of Artificial
  Intelligence, {IAAI} 2021, The Eleventh Symposium on Educational Advances in
  Artificial Intelligence, {EAAI} 2021, Virtual Event, February 2-9, 2021},
  pages 4276--4284, 2021.

\bibitem{DBLP:journals/prl/ManessiR21}
F.~Manessi and A.~Rozza.
\newblock Graph-based neural network models with multiple self-supervised
  auxiliary tasks.
\newblock {\em Pattern Recognit. Lett.}, 148:15--21, 2021.

\bibitem{oliver2018realistic}
A.~Oliver, A.~Odena, C.~Raffel, E.~D. Cubuk, and I.~J. Goodfellow.
\newblock Realistic evaluation of deep semi-supervised learning algorithms.
\newblock In {\em Proceedings of the 32nd International Conference on Neural
  Information Processing Systems}, pages 3239--3250, 2018.

\bibitem{ruder2017overview}
S.~Ruder.
\newblock An overview of multi-task learning in deep neural networks.
\newblock {\em arXiv preprint arXiv:1706.05098}, 2017.

\bibitem{schlichtkrull2018modeling}
M.~Schlichtkrull, T.~N. Kipf, P.~Bloem, R.~Van Den~Berg, I.~Titov, and
  M.~Welling.
\newblock Modeling relational data with graph convolutional networks.
\newblock In {\em European semantic web conference}, pages 593--607. Springer,
  2018.

\bibitem{shi2020masked}
Y.~Shi, Z.~Huang, W.~Wang, H.~Zhong, S.~Feng, and Y.~Sun.
\newblock Masked label prediction: Unified message passing model for
  semi-supervised classification.
\newblock {\em arXiv preprint arXiv:2009.03509}, 2020.

\bibitem{spinelli2020adaptive}
I.~Spinelli, S.~Scardapane, and A.~Uncini.
\newblock Adaptive propagation graph convolutional network.
\newblock {\em IEEE Transactions on Neural Networks and Learning Systems},
  2020.

\bibitem{sun2021scalable}
C.~Sun and G.~Wu.
\newblock Scalable and adaptive graph neural networks with self-label-enhanced
  training.
\newblock {\em arXiv preprint arXiv:2104.09376}, 2021.

\bibitem{sun2020multi}
K.~Sun, Z.~Lin, and Z.~Zhu.
\newblock Multi-stage self-supervised learning for graph convolutional networks
  on graphs with few labeled nodes.
\newblock In {\em Proceedings of the AAAI Conference on Artificial
  Intelligence}, volume~34, pages 5892--5899, 2020.

\bibitem{triguero2015self}
I.~Triguero, S.~Garc{\'\i}a, and F.~Herrera.
\newblock Self-labeled techniques for semi-supervised learning: taxonomy,
  software and empirical study.
\newblock {\em Knowledge and Information systems}, 42(2):245--284, 2015.

\bibitem{trouillon2017knowledge}
T.~Trouillon, C.~R. Dance, J.~Welbl, S.~Riedel, {\'E}.~Gaussier, and
  G.~Bouchard.
\newblock Knowledge graph completion via complex tensor factorization.
\newblock {\em arXiv preprint arXiv:1702.06879}, 2017.

\bibitem{wang2021bag}
Y.~Wang, J.~Jin, W.~Zhang, Y.~Yu, Z.~Zhang, and D.~Wipf.
\newblock Bag of tricks for node classification with graph neural networks.
\newblock {\em arXiv preprint arXiv:2103.13355}, 2021.

\bibitem{wu2019simplifying}
F.~Wu, T.~Zhang, A.~H.~d. Souza~Jr, C.~Fifty, T.~Yu, and K.~Q. Weinberger.
\newblock Simplifying graph convolutional networks.
\newblock {\em arXiv preprint arXiv:1902.07153}, 2019.

\bibitem{wu2021r}
X.~Wu, M.~Jiang, and G.~Liu.
\newblock R-gsn: The relation-based graph similar network for heterogeneous
  graph.
\newblock {\em arXiv preprint arXiv:2103.07877}, 2021.

\bibitem{xu2018representation}
K.~Xu, C.~Li, Y.~Tian, T.~Sonobe, K.-i. Kawarabayashi, and S.~Jegelka.
\newblock Representation learning on graphs with jumping knowledge networks.
\newblock In {\em ICML}, pages 5453--5462. PMLR, 2018.

\bibitem{DBLP:conf/www/0002LS21}
C.~Yang, J.~Liu, and C.~Shi.
\newblock Extract the knowledge of graph neural networks and go beyond it: An
  effective knowledge distillation framework.
\newblock In {\em {WWW} '21: The Web Conference 2021, Virtual Event /
  Ljubljana, Slovenia, April 19-23, 2021}, pages 1227--1237, 2021.

\bibitem{DBLP:conf/icml/YouCWS20}
Y.~You, T.~Chen, Z.~Wang, and Y.~Shen.
\newblock When does self-supervision help graph convolutional networks?
\newblock In {\em Proceedings of the 37th International Conference on Machine
  Learning, {ICML} 2020, 13-18 July 2020, Virtual Event}, pages 10871--10880,
  2020.

\bibitem{yu2020scalable}
L.~Yu, J.~Shen, J.~Li, and A.~Lerer.
\newblock Scalable graph neural networks for heterogeneous graphs.
\newblock {\em arXiv preprint arXiv:2011.09679}, 2020.

\bibitem{yu2020hybrid}
L.~Yu, L.~Sun, B.~Du, C.~Liu, W.~Lv, and H.~Xiong.
\newblock Hybrid micro/macro level convolution for heterogeneous graph
  learning.
\newblock {\em arXiv preprint arXiv:2012.14722}, 2020.

\bibitem{yu2021heterogeneous}
L.~Yu, L.~Sun, B.~Du, C.~Liu, W.~Lv, and H.~Xiong.
\newblock Heterogeneous graph representation learning with relation awareness.
\newblock {\em arXiv preprint arXiv:2105.11122}, 2021.

\bibitem{DBLP:conf/iclr/ZengZSKP20}
H.~Zeng, H.~Zhou, A.~Srivastava, R.~Kannan, and V.~K. Prasanna.
\newblock Graphsaint: Graph sampling based inductive learning method.
\newblock In {\em 8th International Conference on Learning Representations,
  {ICLR} 2020, Addis Ababa, Ethiopia, April 26-30, 2020}, 2020.

\bibitem{zhang2021rod}
W.~Zhang, Y.~Jiang, Y.~Li, Z.~Sheng, Y.~Shen, X.~Miao, L.~Wang, Z.~Yang, and
  B.~Cui.
\newblock Rod: Reception-aware online distillation for sparse graphs.
\newblock In {\em Proceedings of the 27th ACM SIGKDD Conference on Knowledge
  Discovery \& Data Mining}, pages 2232--2242, 2021.

\bibitem{zhang2020reliable}
W.~Zhang, X.~Miao, Y.~Shao, J.~Jiang, L.~Chen, O.~Ruas, and B.~Cui.
\newblock Reliable data distillation on graph convolutional network.
\newblock In {\em Proceedings of the 2020 ACM SIGMOD International Conference
  on Management of Data}, pages 1399--1414, 2020.

\bibitem{zhang2021gmlp}
W.~Zhang, Y.~Shen, Z.~Lin, Y.~Li, X.~Li, W.~Ouyang, Y.~Tao, Z.~Yang, and
  B.~Cui.
\newblock Gmlp: Building scalable and flexible graph neural networks with
  feature-message passing.
\newblock {\em arXiv preprint arXiv:2104.09880}, 2021.

\bibitem{zhang2021evaluating}
W.~Zhang, Z.~Sheng, Y.~Jiang, Y.~Xia, J.~Gao, Z.~Yang, and B.~Cui.
\newblock Evaluating deep graph neural networks.
\newblock {\em arXiv preprint arXiv:2108.00955}, 2021.

\bibitem{distdgl_ai3_2020}
D.~Zheng, C.~Ma, M.~Wang, J.~Zhou, Q.~Su, X.~Song, Q.~Gan, Z.~Zhang, and
  G.~Karypis.
\newblock Distdgl: Distributed graph neural network training for billion-scale
  graphs.
\newblock In {\em 10th {IEEE/ACM} Workshop on Irregular Applications:
  Architectures and Algorithms, {IA3} 2020, Atlanta, GA, USA, November 11,
  2020}, pages 36--44. {IEEE}, 2020.

\bibitem{zhu2021simple}
H.~Zhu and P.~Koniusz.
\newblock Simple spectral graph convolution.
\newblock In {\em International Conference on Learning Representations}, 2021.

\bibitem{aligraph_vldb_2019}
R.~Zhu, K.~Zhao, H.~Yang, W.~Lin, C.~Zhou, B.~Ai, Y.~Li, and J.~Zhou.
\newblock Aligraph: A comprehensive graph neural network platform.
\newblock {\em Proc. VLDB Endow.}, 12(12):2094–2105, Aug. 2019.

\bibitem{zhu2002learnin}
X.~Zhu and Z.~Ghahramani.
\newblock Learning from labeled and unlabeled data with label propagation.
\newblock 2002.

\end{thebibliography}

\appendix

\clearpage
\section{Experiments on ogbn-mag}
\subsection{Compared Baselines}
Ogbn-mag dataset is a heterogeneous graph consists of 1,939,743 nodes and 21,111,007 edges of different types. For comparison, we choose eight baseline methods from the OGB ogbn-mag leaderboard: R-GCN~\cite{schlichtkrull2018modeling}, SIGN~\cite{frasca2020sign}, HGT~\cite{hu2020heterogeneous}, R-GSN~\cite{wu2021r}, HGConv~\cite{yu2020hybrid}, R-HGNN~\cite{yu2021heterogeneous}, NARS~\cite{yu2020scalable}, NARS-SAGN+0-SLE and NARS-SAGN+2-SLE~\cite{sun2021scalable}.  

\subsection{Adapt GAMLP to Heterogeneous Graphs}
In its original design, GAMLP does not support training on heterogeneous graphs.
Here we imitate the model design of NARS to adapt GAMLP to heterogeneous graphs.

First, we sample subgraphs from the original heterogeneous graphs according to relation types and regard the subgraph as a homogeneous graph although it may have different kinds of nodes and edges.
Then, on each subgraph, the propagated features of different steps are generated.
The propagated features of the same propagation step across different subgraphs are aggregated using 1-d convolution.
After that, aggregated features of different steps are fed into our GAMLP to get the final results.
This variant of our GAMLP is called NARS-GAMLP as it mimics the design of NARS.

As ogbn-mag dataset only contains node features for ``paper'' nodes, we here adopt the ComplEx algorithm~\cite{trouillon2017knowledge} to generate features for other nodes.

\subsection{Experiment Results}
We report the validation and test accuracy of our proposed GAMLP and GAMLP+RLU on ogbn-mag dataset in Table~\ref{table.mag_performance}.
%It can be seen from the results that NARS-GAMLP achieves great performance on the heterogeneous graph ogbn-mag, outperforming the performance of the strongest single model baseline NARS-SAGN.
It can be seen from the results that NARS-GAMLP achieves great performance on the heterogeneous graph ogbn-mag, outperforming the performance of the strongest single model baseline NARS.
Equipped with RLU, NARS-GAMLP+RLU further improves the test accuracy, and exceeds the test accuracy of SOTA method NARS-SAGN+2-SLE by a significant margin of 1.50\%.

\begin{table*}[tpb!]
\caption{Test accuracy on ogbn-mag dataset.}
%\vspace{-2mm}
\centering
{
\noindent
\renewcommand{\multirowsetup}{\centering}
\resizebox{0.75\linewidth}{!}{
\begin{tabular}{c|cc}
\toprule
\textbf{Methods} & \textbf{Validation Accuracy} & \textbf{Test Accuracy} \\
\midrule
R-GCN & 40.84$\pm$0.41 & 39.77$\pm$0.46 \\
SIGN & 40.68$\pm$0.10 & 40.46$\pm$0.12 \\
HGT & 49.84$\pm$0.47 & 49.27$\pm$0.61 \\
R-GSN & 51.82$\pm$0.41 & 50.32$\pm$0.37 \\
HGConv & 53.00$\pm$0.18 & 50.45$\pm$0.17 \\
R-HGNN & 53.61$\pm$0.22 & 52.04$\pm$0.26 \\
NARS & 53.72$\pm$0.09 & 52.40$\pm$0.16 \\
NARS-SAGN+0-SLE & 54.12$\pm$0.15 & 52.32$\pm$0.25 \\
NARS-SAGN+2-SLE & 55.91$\pm$0.17 & \underline{54.40$\pm$0.15} \\
%\textbf{NARS-GAMLP} & - & -  \\
\textbf{NARS-GAMLP} & 55.48$\pm$0.08 & 53.96$\pm$0.18  \\
\textbf{NARS-GAMLP+RLU} & 57.02$\pm$0.41 & \textbf{55.90$\pm$0.27}  \\
\bottomrule
\end{tabular}}}
\label{table.mag_performance}
%\vspace{-3mm}
\end{table*}

\clearpage
\section{Detailed Hyperparameters}
We provide the detailed hyperparameter setting on GAMLP+RLU in Table~\ref{table.parameter1},~\ref{table.parameter2} and~\ref{table.parameter3} to help reproduce the results. 
To reproduce the experimental results of GAMLP, just follow the same hyperparameter setting yet only run the first stage.

\begin{table*}[tpb!]
\caption{Detailed hyperparameter setting on OGB datasets.}
%\vspace{-2mm}
\centering
{
\noindent
\renewcommand{\multirowsetup}{\centering}
\resizebox{0.95\linewidth}{!}{
\begin{tabular}{|c|c|c|c|c|c|}
\hline
\textbf{Datasets} & \textbf{attention type} & \textbf{hidden size} & \textbf{num layer in JK} & \textbf{num layer}  & \textbf{activation} \\ \hline
ogb-products   & Recursive & 512  & / & 4 & leaky relu, a=0.2  \\ \hline
ogb-papers100M & JK & 1024  & 4 & 6 & sigmoid \\ \hline
ogb-mag        & JK  & 512  & 4 & 4 & leaky relu, a=0.2\\ \hline
\end{tabular}}}
\label{table.parameter1}
%\vspace{-3mm}
\end{table*}

\begin{table*}[tpb!]
\caption{Detailed hyperparameter setting on OGB datasets.}
%\vspace{-2mm}
\centering
{
\noindent
\renewcommand{\multirowsetup}{\centering}
\resizebox{0.95\linewidth}{!}{
\begin{tabular}{|c|c|c|c|c|c|}
\hline
\textbf{Datasets}   & \textbf{hops} & \textbf{hops for label} & \textbf{input dropout} & \textbf{attention dropout} & \textbf{dropout}           \\ \hline
ogb-products   & 5 & 9 & 0.2 & 0.5 & 0.5 \\ \hline
ogb-papers100M & 6 & 9 & 0 & 0 & 0.5\\ \hline
ogb-mag        & 5 & 3 & 0.1  & 0 & 0.5 \\ \hline
\end{tabular}}}
\label{table.parameter2}
%\vspace{-3mm}
\end{table*}

\begin{table*}[tpb!]
\caption{Detailed hyperparameter setting on OGB datasets.}
%\vspace{-2mm}
\centering
{
\noindent
\renewcommand{\multirowsetup}{\centering}
\resizebox{0.95\linewidth}{!}{
\begin{tabular}{|c|c|c|c|c|c|c|c|}
\hline
\textbf{Datasets}   & \textbf{gamma} & \textbf{threshold} & \textbf{temperature} & \textbf{batch size} & \textbf{stages}         \\ \hline
ogb-products    & 0.1 & 0.85  & 1 & 50000 & 400, 300, 300, 300 \\ \hline
ogb-papers100M & 1 & 0 & 0.001 & 5000 & 100, 150, 150, 150 \\ \hline
ogb-mag        & 10 & 0.4  & 1  & 10000 & 250, 200, 200, 200 \\ \hline
\end{tabular}}}
\label{table.parameter3}
%\vspace{-3mm}
\end{table*}

\end{document}